\documentclass[runningheads,letter,10pt]{llncs}
\usepackage{graphicx}
\usepackage{caption}
\usepackage{amsmath,amssymb}
\usepackage{algorithmic,algorithm}
\usepackage{multicol,multirow}
\usepackage{color}
\usepackage{navigator}

\newif\ifBibtex\Bibtexfalse
\newif\ifDraft\Draftfalse

\newcommand{\figref}[1]{Fig.\ref{#1}}
\newcommand{\tabref}[1]{Table \ref{#1}}
\newcommand{\algref}[1]{Algorithm \ref{#1}}

\ifDraft
\newcommand{\rewrite}[1]{\textcolor{red}{#1}}
\newcommand{\relocate}[1]{\textcolor{blue}{#1}}

\else
\newcommand{\rewrite}[1]{{#1}}
\newcommand{\relocate}[1]{{#1}}

\fi
 
\begin{document}
\title{Weak-supervision for Deep Representation Learning under Class Imbalance}
\author{Shin Ando\inst{1}
}
\authorrunning{Shin Ando}
\institute{Tokyo University of Science, \\ Chiyoda-ku, Tokyo, Japan, 1020071 \\ 
 \email{ando@rs.tus.ac.jp}\\
 \url{http://www.rs.tus.ac.jp/ando} 
}
\maketitle              
\begin{abstract}
Class imbalance is a pervasive issue among classification models including deep learning, whose capacity to extract task-specific features is affected in imbalanced settings. However, {the challenges of handling imbalance among a large number of classes, commonly addressed by deep learning, have not received a significant amount of attention in previous studies.} In this paper, we propose an extension of the {deep} {over-sampling} framework, to exploit automatically-generated abstract-labels, i.e., a type of side-information used in weak-label learning, to enhance deep representation learning against class imbalance. We attempt to exploit the labels to \rewrite{guide the deep representation of instances towards different subspaces, to induce a \emph{soft}-separation of inherent subtasks of the classification problem.} Our empirical study shows that the proposed framework achieves a substantial improvement on image classification benchmarks with imbalanced among large and small numbers of classes.
\end{abstract}
\section{Introduction}
The advances of deep learning models that enable automatic extraction of discriminative features from a massive amount of labeled data have eased the burden of hand-engineering them in many classification applications. {The preparation of ground-truth labels, in turn, has become critical for those applications, and in cases where its cost is too steep, techniques to exploit additional information, such as transfer learning and weakly-supervised learning, are employed.}
 
The problem of class imbalance can occur in cases where the preparation of labeled data is difficult for specific classes. The imbalanced settings can typically deteriorate the retrieval measures for the classes of \emph{minority} \cite{Branco:2016:SPM:2966278.2907070,Krawczyk2016}, as well as the representation learning of deep neural nets \cite{10.1007/978-3-319-71249-9_46}.

On the topic of class imbalance, a large portion of the studies have focused on binary cases and multi-class cases with less than ten classes. However, deep learning models commonly address problems with a much larger number of classes, at which the impact is much more difficult {to handle}.
In this paper, we attempt to leverage a type of weak-labels to address class imbalance over a large number, e.g., up to one hundred, of classes. 
Weak-labels are side-information used in weakly-supervised learning to complement a limited amount of labeled data. They are generated automatically or by inexpensive means such as labeling functions \cite{delbp18-ratner} 
 and crowdsourcing, and usually of low-quality or abstract-level \cite{doi:10.1093/nsr/nwx106}.

In this paper, we consider external knowledge in the form of abstract-labels assigned to every training instances, providing a categorization of the original classes. For example, training instances with classes: \{\emph{airplane, automobile, bird, cat, deer, dog, frog, horse, ship, truck}\}, may be categorized with abstract-labels: \{\emph{animals, vehicles}\} using a set of rules: \{\emph{bird, cat, deer, dog, fish, horse}\}$\to$\emph{animals}, \{\emph{airplane, automobile, ship, truck}\}$\to$\emph{vehicles}.
We assume that, as in this example, each class is categorized by only one label. 

{Such categorization can contain relevant information regarding the hierarchical structure of the classes, and provide useful guidance for learning the structure of the deep representation to counter the effect class imbalance which may be over- or under-estimation of class boundaries.} 
{Our intuition to exploit such abstract-labels, to this end, is to acquire deep representation which induce the separation of the subtasks, i.e., discriminating among a subset of classes categorized to each label.} We implement a framework to associate an independent subspace of the deep features to each label and guide the instances towards the targets projected onto the corresponding subspaces.

We build on the framework of Deep Over-sampling (DOS) \cite{10.1007/978-3-319-71249-9_46} which drew inspiration from the classic synthetic minority over-sampling \cite{Chawla:2002:SSM:1622407.1622416}
{that re-balances the class distribution by augmenting synthetic minority-class instances sampled from the neighborhood of existing ones into the training data.}
DOS integrates re-sampling into deep learning, by implementing an additional back-propagation for the output of the embedding layers to directly guide its representation learning.

The caveat on the weakly-supervised learning is that the weak-label may not always be of ideal granularity or hierarchical structure. That is, \rewrite{enforcing instances onto orthogonal subspaces that reflect the abstract-label categorization may not be entirely beneficial.}
\rewrite{Alternatively, we attempt to induce a \emph{soft}-separation of subtasks
using the gradient of squared-sum error to gradually separate the representations of different labels in terms of cosine distance.}  
\rewrite{The proposed framework can also benefit from the multi-task learning framework, where the net parameters are simultaneously trained based on the standard class prediction, which can counteract the detrimental aspects of weak-supervision.}

\section{Background}\label{sec:related_work}

\subsection{Class imbalance}
Class imbalance is a practical issue, where a large discrepancy in the number of samples among classes causes the learning algorithm to over-generalize for the classes in the \emph{majority}. Its  effect on the retrieval measures for the \emph{minority}  classes, which is of the primary interest for many applications, is critical.

The typical approaches to counter class imbalance include re-sampling, instance-weighting, and cost-sensitive learning. The re-sampling approach directly addresses the imbalance by over- or under-sampling on the training data. 
Synthetic Minority Over-sampling (SMOTE) is a popular over-sampling method, which worked successfully with many traditional classification models. 

It was demonstrated in \cite{10.1007/978-3-319-71249-9_46} that use of SMOTE on the deep representation acquired by a convolutional neural net (CNN) under the effect of imbalance do not yield as much merit as it does with hand-engineered features. To address this issue, they proposed Deep Over-sampling (DOS), which implemented the sampling of the minority class into the deep learning process. The synthetic samples were used as supervising targets for the representation learning, which provided an additional feed-back to the embedding layers of the CNN to improve the in-class and inter-class separation among deep representation.

With regards to the number of classes, the previous studies on class imbalance have mainly focused on binary classification and cases with a small number of classes.
Imbalance among a large number of classes, meanwhile, has received limited attention. 
In recent surveys, the class imbalance problems been categorized between binary or multi-class and only few multi-class methods have addressed more than ten classes \cite{He:2009:LID:1591901.1592322,Krawczyk2016,Fernandez2017}.

\subsection{Weakly-supervised Learning}
Supervised learning, deep learning especially, requires a large amount of labeled data, which can be costly in some applications. Weakly-supervised learning exploits various side-information from crowdsourcing, heuristic labeling, external knowledge base, etc., to \rewrite{achieve better} performances \cite{doi:10.1093/nsr/nwx106}.
Techniques, such as semi-supervised learning, multi-instance learning, and learning with noisy-labels are employed to address various conditions of weak-labels, including coverage, granularity, or accuracy.

Heuristic labeling, as opposed to hand-curated annotations, require little cost for assigning weak-labels to the training data. In \cite{Ratner:2017:SRT:3173074.3173077}, labeling functions written by domain experts were used to generate labels specifying the hierarchy of sub-tasks in image and document classification. The relation between the weak-labels and the ground-truth labels were captured using an exponential family generative model which is integrated into training the discriminative model.

As mentioned in the previous section, we consider abstract-labels that categorize the original classes and is relevant to the task at hand, such that dividing subsets of classes with abstract-labels can reduce the complexity of the classification problem.
\rewrite{But as with other weakly-supervised learning scenarios, we take into account that hierarchical structure of the label may not be ideal and possibly introduce added perplexity.}
 
\subsection{Preliminary Results}\label{sec:preliminaries}
To demonstrate the motivation of our study, we conducted an experiment with artificially imbalanced settings%
. We modify the CIFAR-10 image benchmark \cite{citeulike:7491128} by removing 80\% of the samples from selected classes. {Then, we trained a CNN with VGG16 layers of pre-trained weights \cite{DBLP:journals/corr/SimonyanZ14a} and two fully-connected layers of randomly initialized weights}.

Following imbalanced settings were compared in our analysis: 1) maintaining 100\% of samples of all classes (Full Data), 2) maintaining 100\% of samples in six classes and removing 80\% of samples from four classes (Imbalanced), 3) removing 80\% of samples from all classes (Balanced). In setting (2), we refer to the four classes from which we removed the samples as \emph{minority} classes and the rest as \emph{majority} classes.
For evaluation, each experiment was repeated ten times from different initial parameters. The four minority classes and the removed samples were chosen randomly for each repetition.

In \tabref{tab:accuracy_prelim}, we compare the accuracies, which is standard when using the full data. The drop-off from (1) to (2) indicates the impact of the class imbalance. We note that the parameters of VGG16 net is trained on a balanced dataset, thus reduces  the impact of imbalance, and that the discrepancy is much larger if the entire parameters are obtained from the imbalanced training set.
Furtheremore, the significant difference between (2) and (3) suggests that the imbalance in sample sizes can be detrimental even if the number of samples is larger in total.
\begin{table}
\begin{minipage}[t]{.5\textwidth}
\caption{Accuracies}\label{tab:accuracy_prelim}
\centerline{
\begin{tabular}{ll}
\hline
(1) Full Data & $0.87\pm0.0053$ \\
(2) Imbalanced Reduction &$0.84\pm0.013$\\
(3) Balanced Reduction &$0.85\pm0.011$\\
\hline
\end{tabular}
}
\end{minipage}
\begin{minipage}[t]{.5\textwidth}
\caption{Class-wise precision/recall}\label{tab:precision/recall_prelim}
\centerline{
\begin{tabular}{ccl}
\hline
\multirow{2}{*}{Majority} & precision & $0.77\pm0.017$ \\
& recall & $0.92\pm0.0089$ \\
\hline
\multirow{2}{*}{Minority} & precision & $0.92\pm0.013$ \\ 
& recall &$0.79\pm0.025$ \\
\hline
\multirow{2}{*}{Balanced} & precision &$0.86\pm0.0093$\\
& recall & $0.85\pm0.011$\\
\hline
\end{tabular}
}
\end{minipage}
\end{table}

For further investigation, we evaluated the class-wise precision and recall in (2) and (3) as shown in \tabref{tab:precision/recall_prelim}. The first two rows show the measurements on the majority and the minority classes from (2) and the third row shows the measurements from (3). The impact of class imbalance is shown strongly in the recall of the minority classes and the precision of the majority classes, which are significantly worse \rewrite{than in (3), with smaller but balanced number of samples.}

From the preliminary results, we found that class imbalance can affect the discriminative power over the classes in the majority as well as the minority, and addressing it can be more crucial than the preparation of majority class samples.

\section{Deep Subspace Sampling}\label{sec:deep_subspace_over-sampling}

\subsection{Basic Definitions}
We denote the layers of the CNN as two groups: the embedding layers and the classification layers. The former projects the input onto the deep feature space, and the latter makes the class prediction from the feature vectors. We denote the function of the embedding layers as $f:\Phi\to\mathbb{R}^d$, where $\Phi$ is the domain of input data. The function of the classification layers, whose output is the vector of class probabilities over $k$ classes, is denoted as $g:\mathbb{R}^d\to[0:1]^{n}$.

The training data is a set of input/output pairs denoted by ${\mathcal X}={(x^{(i)},y^{(i)})}_{i=1}^n$, where $x^{(i)}\in\Phi$ and the output $y^{(i)}$ takes a values from a set of classes ${\mathcal C}=\{c_1,\ldots,c_k\}$.
Additionally, weak-supervision is provided by a deterministic labeling function $\Lambda:{\mathcal C}\to{\mathcal L}$, where ${\mathcal L}=\{\lambda_1,\ldots,\lambda_l\}$ denotes a set of abstract-labels. We denote the abstract-label of the $i^{\text{th}}$ data by $z^{(i)}$, i.e., $z^{(i)}=\Lambda(y^{(i)})$.%
 
Let ${\mathcal V=\{f(x^{(i)})\}}$ denote the set of projections of the input by the tentative embedding function $f$. We define the subset of ${\mathcal V}$  belonging to class $c$ as ${\mathcal V}({c})=\{f(x^{(i)}):y^{(i)}=c\}$. The subset of projections with abstract-label $\lambda$ is denoted by ${\mathcal V}_{\lambda}=\{f(x^{(i)}:z^{(i)}=\lambda\}$. 

\subsection{Deep Over-sampling}
Following the DOS framework \cite{10.1007/978-3-319-71249-9_46}, we implement an in-class neighborhood sampling over the deep projections to generate a multi-task learning training set.
For each $x^{(i)}$ in $\mathcal{X}$, 
a subset of $m$  in-class neighbors ${\mathcal N}$ defined as 
\begin{equation}
{\mathcal N}_m\left(x^{(i)}\right)=\mathop{\arg\min}\limits_{\begin{matrix}\mathcal{S}\subset{\mathcal V}\left(y^{(i)}\right)\\\#(S)=m\end{matrix}}
\sum\limits_{v\in{\mathcal S}}
\|f(x^{(i)})-v\|^2
\end{equation}
is selected.
We generate each instance of multi-task training set is as a tuple $(x,y,{\mathcal N}(x),{\mathbf w})$ where $(x,y)$ is the original input/output, ${\mathcal N(x)}$ its neighbors, and 
random weights ${\mathbf w}=(w_1,\ldots,w_k)$ which sums to 1, i.e., $\sum_i w_i=1$.

By randomly sampling the weights ${\mathbf w}$, we can sample \emph{different} numbers of tuples from a common original input/output $(x_i,y_i)$, and we generate the training set ${\mathcal X}'=\{(x^{(i)},y^{(i)},{\mathcal N},{\mathbf w})\}_{j=1}^{n'}$ such that the sample sizes are balanced among all classes. 
 
The network architecture for multi-task learning includes two outputs: one at the classification layers and the other at the embedding layers. The back propagation for the former output is implemented with a standard, cross-entropy loss on class prediction. For the latter, a squared-sum loss is defined as follows.
\begin{equation}
l(x,y,{\mathcal N},{\mathbf w})=\sum_{v\in{\mathcal{N}}} w_i \|f(x)-v\|^2
\label{eq:loss}
\end{equation}
\eqref{eq:loss} is minimized when the deep representation of the original input is at an interpolation of the neighbors. This loss thus sets a target for the embedding function, which guides the representation towards the class-mean, as the local means distribute closer to the class mean than the original samples. 
The DOS framework makes up for induces smaller in-class variance by iteratively updating the representation with this process and

\subsection{Subspace Selection}
The proposed framework builds on DOS to exploit abstract-labels, by guiding instances of different labels \rewrite{towards} independent subspaces. This section describes two approaches for selecting such subspaces: \emph{fixed subspace allocation} and \emph{supervised subspace selection}. 

Let ${\mathcal U}_i$ denote the subspace corresponding to the label $\lambda_i$ and ${\mathcal B}_i$ its basis. For simplicity, we define the dimensionality of all subspaces to be $p$, such that $p\times{l}<d$.
The fixed subspace allocation simply assigns a subset of variables to each label, defining, in turn, a subspace where all complementary variables are zero. 
The basis ${\mathcal B}_i=\{(b_{i1},\ldots,b_{id})\}_{i=1}^{p}$ is defined such that 
$b_{ij}=1$ for $j\in[1+p(i-1):p+p(i-1)]$ and $b_{ij}=0$ for all other $j\in[1:d]$. 

In the supervised subspace selection, we attempt to find subspaces that preserve the discriminative information relevant to the subtask of each label. To that end, we adopt a supervised dimensionality reduction method, such as linear discriminant analysis, as a function into the following process. The function takes ${\mathcal V}(\lambda)$ as an input and returns a primary component ${\mathbf b}$, such as the first eigenvector, that maximizes its classification objective for a subset of classes $\{c:\Lambda(c)=\lambda\}$.

A brief description of the supervised subspace selection is given as follows. 
\begin{enumerate}
  \item Initialize the basis $\{{\mathcal B}_1,\ldots,{\mathcal B}_k\}\leftarrow\{\emptyset,\ldots,\emptyset\}$
  \item Set ${\mathcal V}^1={\mathcal V}$. 
  \item For $t=1,\ldots,q$.
  \item Select a label $\lambda_i\in{\mathcal L}$
  \item Compute a primary component ${\mathbf b}$ by supervised dimensionality reduction over ${\mathcal V}^t(\lambda_i)$ 

  \item Update ${\mathcal B}_i\leftarrow{\mathcal B}_i\cup{\mathbf b}$
  \item Update ${\mathcal V}^{t+1}$ with a projection of ${\mathcal V}^{t}$ such that ${\mathcal V}^{t+1}\perp{\mathbf b}$
  \item Back to step 3
\end{enumerate}
In essence, this process iteratively allocates a discriminative component to a subspace and removes it from the the original or residual representation.
In step 4, the label $\lambda_i$ is selected randomly and evenly so that each label is chosen $p$ times over the entire repetition, thus $q=l\times{p}$ at step 3.
The steps 3-8 are not necessarily repeated until ${\mathcal V}^{q+1}=\emptyset$ since $q<d$ in general. In such a case, the basis of the residual subspace ${\mathcal V}^{q+1}$ can simply be appended to the basis of all subspaces. 

\figref{fig:visualization} illustrates our intuition for guiding the deep representation towards independent subspaces.
\rewrite{The blue markers indicate the tentative vector representation in the deep feature space. Each instance comes from different classes, as indicated by callout texts and the shapes of the markers. Let us assume that the abstract-labels of the \emph{airplane} and \emph{ship} classes, indicated by solid markers, is label 1 and that of \emph{fish} and \emph{bird} classes, indicated by circled markers, is label 2. Let $x$- and $y$-axes represent the subspaces associated with labels 1 and 2, respectively.}

{The red markers indicate the synthetic targets generated from in-class neighbors. The targets for label 1-classes are generated on subspace 1 and those for label 2-classes are generated on subspace 2.} 
{Descending the gradient of the squared-sum error, their representations are updated to be closer to the red markers. The green markers indicate the updated vector representation.}

\rewrite{By guiding representation towards these subspaces, we aim to induce a structure where the subtask can be addressed more independently, even if their representations are not strictly orthogonal, as the cosine distances among the classes assigned to different labels can grow larger.} 

\subsection{Multi-task learning}
\begin{figure}[tb]
\begin{minipage}[t]{0.5\textwidth}
\caption{DS3 Framework}\label{fig:SOS_framework}
\centerline{
\includegraphics[height=4.5cm]{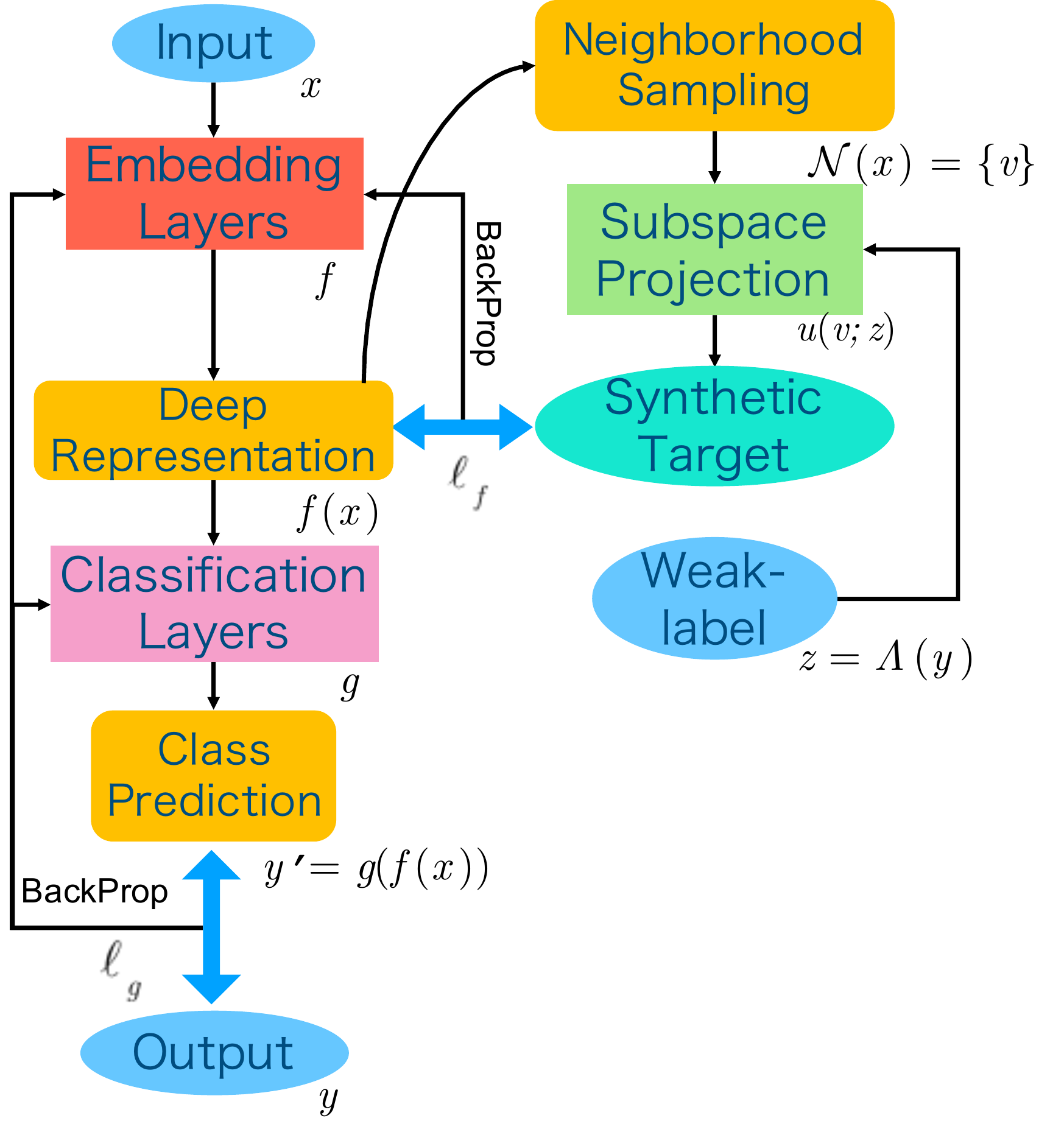}
}
\end{minipage}
\begin{minipage}[t]{0.5\textwidth}
\caption{Subspace Projections}\label{fig:visualization}
\centerline{
\includegraphics[height=4.5cm]{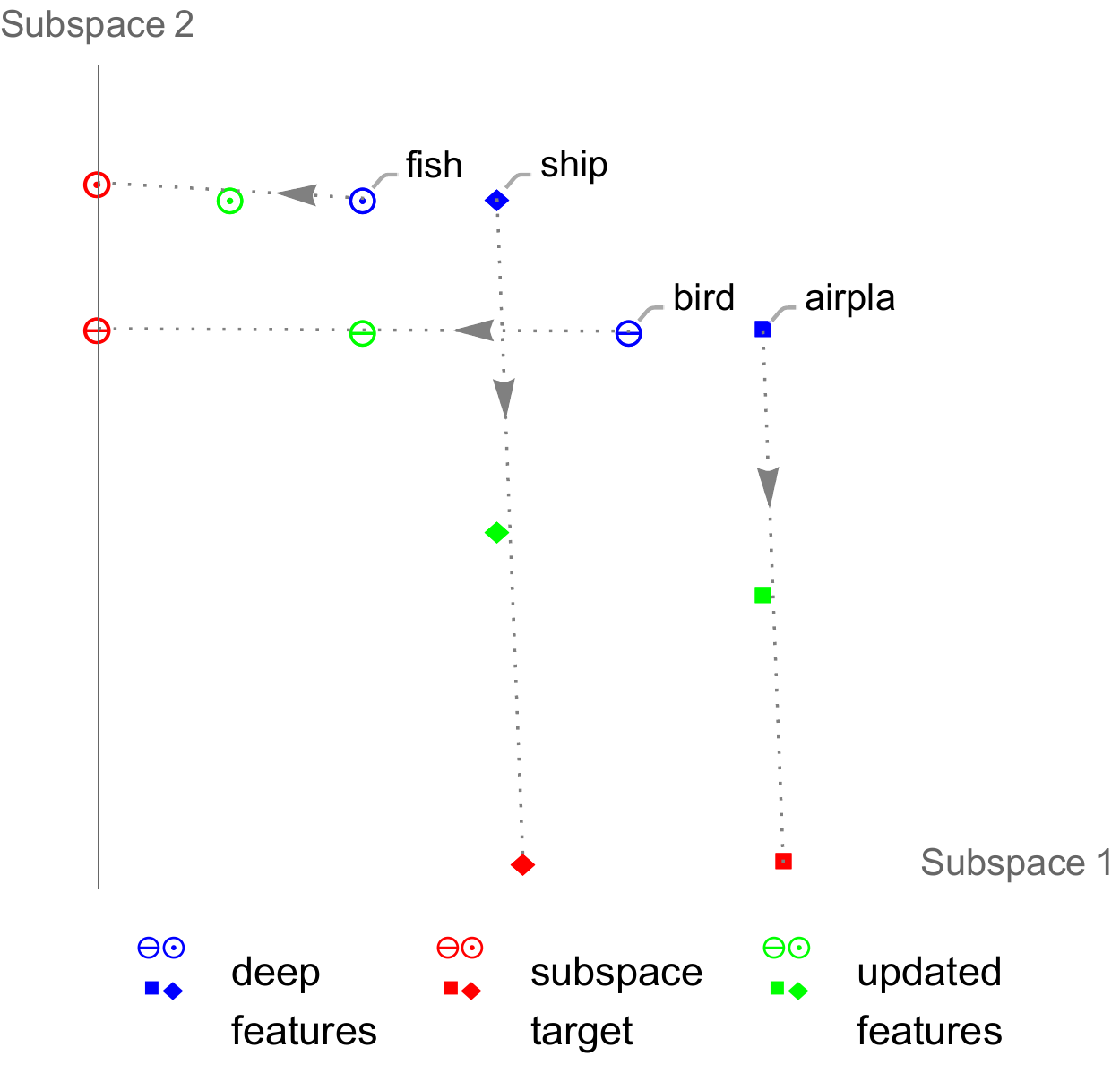}
}
\end{minipage}
\end{figure}

This section describes the multi-task learning framework which integrates representation learning using the subspaces described in the previous section.
The overview of the framework is illustrated in \figref{fig:SOS_framework}.

On the left side of the figure, the basic architecture of a deep neural net comprised of the embedding layers and the classification layers is shown. Their outputs are the deep representation and the class prediction, respectively.
The input and the output makes up the original training instance, and a single-task learning can be conducted with a back-propagation from the prediction error.

on the right side of \figref{fig:visualization}, the components related to representation learning with re-sampling and weak-supervision are shown. The synthetic target and the weak-label are additionally included in the training instance for multi-task learning. 
Another back-propagation explicitly for the embedding layers is prompted by the difference between the synthetic target and the deep representation.
\relocate{The proposed framework is referred to as Deep SubSpace Sampling (DS3).}

The training instance for multi-task learning is a tuple $\left(x^{(i)},y^{(i)},{\mathbf w},{\mathcal N}(x^{(i)}),z^{(i)}\right)$, where $z^{(i)}=\Lambda(y^{(i)})$.
The loss function for the classification function $g$ is 
the standard cross-entropy loss
\begin{equation}
\ell_g(x,y,{\mathcal N},{\mathbf w},z)=H(g(v),y)
\label{eq:loss3}
\end{equation}

Additionally, the loss function for the embedding function $f$ is defined as 
a weighted mean squared-sum error  
\begin{equation}
\ell_f(x,y,{\mathbf w},{\mathcal N},z_j)=\alpha\sum_{v\in{\mathcal{N}}}^k w_i \|f(x)-u(v;z_j)\|^2
\label{eq:loss2}
\end{equation}
where $u(v;\lambda_j)$ denote the projection of deep feature vector $v$ onto the subspace ${\mathcal U}_j$ and $\alpha$ is a trade-off coefficient against the first loss in \eqref{eq:loss3}.

Recalling that ${\mathbf w}$ is a randomized weight vector, one of the merit from \eqref{eq:loss2} is inducing robustness to avoid overfitting, similar to that of adding noise to the output of different layers. Another merit comes from guiding the instances toward an interpolation of the in-class neighbors and closer to the class mean. Subsequently, it can increase the inter-class discrepancies in the deep feature space.

\rewrite{The two back-propagations based on the above two loss functions constitutes a multi-task learning of a standard classification learning and an explicitly supervised representation learning.} Note that while the propagation from \eqref{eq:loss3} updates parameters of all layers, \eqref{eq:loss2} only affects those of the embedding layers.

\subsection{Subspace Sampling Algorithm}
The Deep Subspace Sampling framework combines the merits of over-sampling and  explicitly supervised representation learning with weak-supervision by abstract-labels.
Its multi-task learning framework allows for (a) augmenting training set with synthetic projections of the minority class samples, (b) inducing robustness with randomized targets, and (c) \relocate{separating subspaces to acquire discriminative information for different subtasks.}

The overview of the algorithm is shown in \algref{alg:DS3}.
\begin{algorithm}[h]
\caption{Deep Subspace Sampling algorithm}\label{alg:DS3}
\begin{algorithmic}[1]
\STATE {\bf Input}: Training set ${\mathcal X}$, class-wise over-sampling size $\{r_j\}_{j=1}^k$, abstract-labels $\{\lambda_j\}_{j=1}^l$, \# of training rounds $T$
\STATE {\bf Output}:  A trained CNN 
\STATE {\bf function} SubspaceSelect: subspace selection method
\STATE {\bf Method}:
\STATE Initialize CNN by single task learning with ${\mathcal X}$
\FOR{$t=1,\ldots,T$}
\STATE Compute projections ${\mathcal V}$ from ${\mathcal X}$
\FOR{$j=1,\ldots,l$}
\STATE $U_j\leftarrow\text{SubspaceSelect}({\mathcal V}(z_j))$
\ENDFOR
\STATE ${\mathcal X'}=\emptyset$
\FOR{$i=1,\ldots,n$}
\STATE Set resampling size $R {\leftarrow} r_j:y^{(i)}=c_j$
\FOR{$j=1,\ldots,R$}
\STATE ${\mathcal N}\leftarrow$NeighboorhoodSampling$(x^{(i)},k)$
\STATE Generate random weight ${\mathbf w}$
\STATE ${\mathcal X'}\leftarrow{\mathcal X'}\cup\{(x,y,{\mathcal N},{\mathbf w},z)\}$
\ENDFOR
\ENDFOR
\STATE Update CNN by multi-task learning with ${\mathcal X}'$ 
\ENDFOR
\end{algorithmic}
\end{algorithm}
The initial embedding and classifier functions are obtained by a standard training a CNN with the original, imbalanced data, at line 5.
\relocate{The basis for each subspace is updated after each update of the CNN,} 
at lines 8-10, assuming that the subspace selection method is supervised selection.
\rewrite{If the subspace selection method is fixed allocation, it can be executed at initialization  at line 6, i.e., before the start of the outer loop.}
{The synthetic targets are re-computed at each iteration as in-class neighbors are updated with the  deep representation.} 

{The trade-off coefficient $\alpha$ is an important hyper-parameter which controls} 
\rewrite{the speed of the descent towards orthogonal representation, which could be detrimental if it is too strong, relative to the classification learning.}
Its value was selected empirically as described in the next section.
\rewrite{The computationally intensive operations in this process, outside of deep learning, are the dimensionality reduction at line 9 and the neighbor search at line 13.  A practical run time analysis is also provided in the next section.}  

\section{Empirical Results}\label{sec:emprical_study}
The empirical study is organized in three parts: (1) comparison between the two subspace selection methods, (2) sensitivity analysis on \rewrite{essential} parameters, and (3) comparative analysis with baseline methods%
. 

\subsection{Datasets}
The four image classification benchmarks were used in this experiment are: CIFAR-10/100 \cite{citeulike:7491128}, STL-10 \cite{coates2011analysis}, and SVHN \cite{37648}.
The properties of the datasets are summarized in \tabref{tab:dataset_summary}.
All results are reported on the default test split.
\begin{table}[tb]
\caption{Summary of Datasets}\label{tab:dataset_summary}
\centerline{\small
\begin{tabular}{cccc}
\hline
Dataset&\#channels$\times$size & \#images per class (train/test) & \#classes/abstract-labels\\
\hline
CIFAR-10 & $3\times32\times32$ & 5000/1000 & 10/2 \\
SVHN & $3\times32\times32$ & 7000/2000 &10/2 \\
STL-10 &  $3\times96\times96$ & 500/800 & 10/2 \\
CIFAR-100 & $3\times32\times32$ & 500/100 & 100/8 \\
\hline
\end{tabular}
}
\end{table}

For the CIFAR-10 and STL-10 datasets, the abstract-labels \{animals,vehicles\} were assigned by semantic rules\footnote{CIFAR-10: \{airplane, automobile, ship, truck\}$\to$vehicle, \{bird, cat, deer, dog, frog, horse\}$\to$animal, STL-10: \{airplane, car, ship, truck\}$\to$vehicle, \{bird, cat, deer, dog, horse, monkey\}$\to$animal}.
For the CIFAR-100 dataset, the twenty super-classes in the original dataset were used as abstract-labels. 
For SVHN, labels \{odd, even\} were assigned according to the class digits. \rewrite{With this set of labels, we measure the effect of abstract-labels that do not provide relevant information for the task.} 

The imbalanced settings were set up by randomly selecting 50\% of the classes from each label to be the minority classes and removing 80\% of their samples also at random.
\subsubsection{Settings and Evaluation}
The same CNN architecture is trained by a standard single-task learning, the DOS framework, and the DS3 framework. The first two models are used to provide the baselines for a comparative analysis.
\rewrite{For the CIFAR-10/100 and STL-10 datasets, we employed the architecture of VGG16 \cite{DBLP:journals/corr/SimonyanZ14a} joined to two fully-connected layers of randomly initialized weights, or C64-C128-C256-C512-C512-F$p$-F$n$}. $p$ is the dimensionality of the deep representation vectors, which was set to $1000$, and
$n$ is the number of classes. 
For the SVHN dataset, we employed the architecture used in \cite{10.1007/978-3-319-71249-9_46}, two convolutional layers with 6 and 16 filters, respectively joined to two fully-connected layers, or C6-C16-F400-F120. 

\rewrite{The number of training rounds $T$ was set to 8, after empirical analysis described in a later subsection}.
The neighborhood sampling size $m$ is set to five for the DOS and the DS3 framework.
Deep learning was conducted on NVIDIA TITAN V graphic card with 2560 cores and 12 GB global memory. 
For evaluation, we measured the overall accuracy and three retrieval measures: precision, recall, and F1-score averaged over the majority and the minority classes, respectively. We include the accuracy for comparison against \emph{full-data} performances.

\subsection{Subspace Selection}
We first compare between the two subspace selection methods: the supervised selection and fixed allocation. \tabref{tab:subspacemethods} summarizes their retrieval measures on artificially imbalanced CIFAR-10 and STL-10 datasets.

\begin{table}[tb]
\begin{minipage}[t]{0.5\textwidth}
\caption{Subspace selection comparison}\label{tab:subspacemethods}
\centerline{
\begin{tabular}{ccccc}
\hline
&&{Fixed}&{Supervised}\\
&&(min/maj)&(min/maj)\\
\hline
\multirow{3}{*}{CIFAR-10}
&Pr &0.919/0.809&0.928/0.801\\
&Re &0.792/0.914&0.778/0.920\\
&F1 &0.851/0.857&0.845/0.854\\
\hline
\multirow{3}{*}{STL-10}
&Pr &0.882/0.666&0.886/0.666\\
&Re &0.576/0.907&0.571/0.903\\
&F1 &0.694/0.772&0.692/0.766\\
\hline
\end{tabular}
}
\end{minipage}
\begin{minipage}[t]{0.5\textwidth}
\captionsetup{type=figure}
\caption{Convergence of DS3 on CIFAR-100}\label{fig:convergence}
\centerline{
\includegraphics[width=5.5cm]{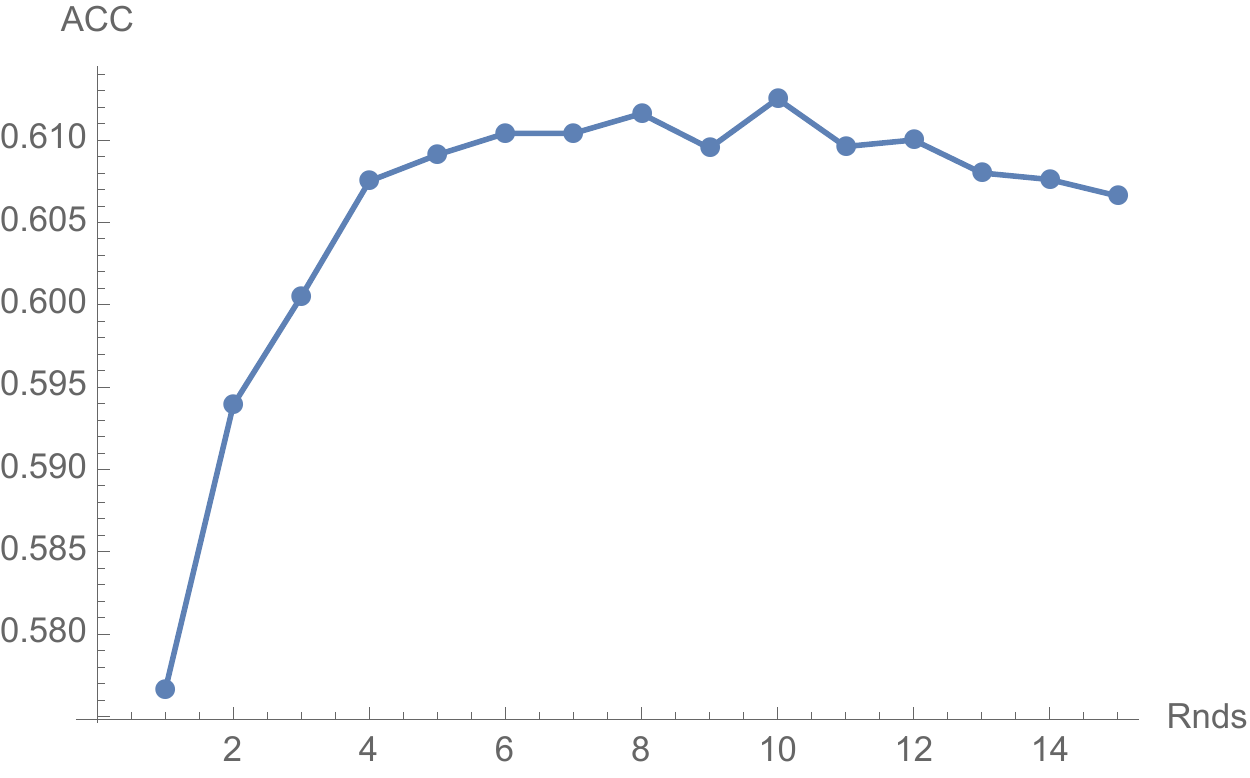}
}
\end{minipage}
\end{table}
The third and the fourth columns show the measurements for the Fixed allocation and the Supervised selection, respectively. The precision (Pr), recall (Re), and F1 are averaged over the minority (min) and the majority (maj) classes. The averages are taken over ten repetitions.

From the results on F1 and accuracy, the \rewrite{fixed allocation yielded slightly better average over the supervised selection and also smaller deviations.}
In addition, we observed that fixed allocation showed small but significant advantage in the average recall of the minority classes.
Overall, we found fixed allocation to be a preferable approach with regards to the retrieval performances and the computational complexity. In the following, we report the results of DS3 with fixed subspace allocation.

\subsection{Sensitivity and Convergence Analysis}
Next, we evaluated the sensitivity of the DS3 framework regarding the trade-off weight $\alpha$ with a grid search over five values: $\{0.0025,0.005,0.01,0.02,0.04\}$  in ten repetitions over imbalanced CIFAR-10 and STL-10.

We observed that all measures show similar tendencies, and only report the average accuracies for brevity. The results are summarized in \tabref{tab:alpha}.
On both datasets, the accuracies were reasonably robust over the tested values. \rewrite{We report the results in the following section for $\alpha=0.01$}.
\begin{table}[tb]
\caption{Sensitivity Analysis}\label{tab:alpha}
\centerline{
\begin{tabular}{ccccccc}
\hline
$\alpha$ & 0.0025 & 0.005 & 0.01 & 0.02 & 0.04\\
\hline
CIFAR-10 & 0.852&0.853&0.854&0.853&0.852 \\
STL-10 & 0.748&0.757&0.757&0.753&0.752 \\
\hline
\end{tabular}
}
\end{table}

Additionally, we empirically analyzed the convergence of accuracy. \figref{fig:convergence} shows a typical run of DS3 on CIFAR-100.
\rewrite{The $x$- and $y$-axes indicate the iteration and the accuracy, respectively.}
\rewrite{We observed that the performance can start to decline after $10$ rounds, and selected $T=8$ for the following experiment.}

\subsection{Comparative Analysis}
In the comparative analysis, we report the average F1 scores on the minority and majority classes for the proposed and the baseline methods, as all retrieval measures showed similar tendency over the four datasets. The accuracies are also reported for comparison with standard full data training. The results are summarized in \tabref{tab:summary_imbalance}. 
\begin{table}[tb]
\caption{Comparative Analysis (Accuracy/minority F1/majority F1)}\label{tab:summary_imbalance}
\centerline{
\begin{tabular}{ccccc}
\hline
& CNN & DOS & DS3 & CNN (Full) \\
\hline
CIFAR-10 & 0.829/0.826/0.835 & 0.853/0.851/0.854 & 0.854/0.851/0.857 & 0.857\\
SVHN & 0.520/0.462/0.648 & 0.758/0.746/0.763 & {0.755}/0.744/0.762 & 0.850 \\ 
STL-10 & 0.723/0.667/0.757 & 0.741/0.694/0.772 & 0.757/0.702/0.786 & 0.801 \\
CIFAR-100 & 0.558/0.355/0.629 & 0.594/0.355/0.629 & 0.612/0.390/0.635 & 0.638\\
\hline
\end{tabular}
}
\end{table}

In each of the first three columns, three measurements: accuracy, average F1 over minority classes, and average F1 over majority classes, of the baseline methods and DS3. 
The last column, CNN (Full), shows the accuracy of the baseline CNN trained with the full set of original data. 
The difference between the first and the last column thus provides a reference to the impact of class imbalance. We observed that DOS was able to make up for a large portion of the impact on CIFAR-10 and SVHN. Meanwhile, the margin of improvement by DOS is relatively much smaller for CIFAR-100 and STL-10.

DOS showed a substantial setback on CIFAR-100, which may be attributed to the large number of classes. \rewrite{That is, the usefulness of synthetic samples may possibly diminish with a larger number of minority classes to supplement.}
With regards to STL-10, the difficulty may be attributed to the smaller number of training samples.

The DS3 exhibited large advantages over DOS for CIFAR-100 and STL-10 and a slight improvement with CIFAR-10. In the case of CIFAR-10, the performance of DOS is very close to that of CNN trained with full data and have little room for improvement.

In the case of SVHN, improvements by DS3 were not expected as the abstract-labels did not provide information relevant to the task. The results showed that the difference between DS3 and DOS were not significant, thus DS3 did not yield negative effects from irrelevant labels.

To summarize, the DS3 framework was able to make larger improvements over the baseline DOS in more difficult problems, i.e., with a larger number of classes or with fewer samples.
In the other problems, it achieved better or nearly equivalent performances
as the baseline even in cases where the abstract-labels were not relevant.
Finally, the run time of DS3 compared to the standard training of CNN ranged from 24\% to 41\% increases among the four datasets. While the increase in computational time is inevitable due to the neighborhood sampling, the trade-off is justifiable in cases of substantial imbalance.
 
\section{Conclusion}\label{sec:conclusion}
We proposed the Deep Subspace Sampling framework for utilizing automatically-generated abstract-labels in deep representation learning to enhance its robustness against class imbalance. 
\rewrite{It exploits the abstract labels to learn deep representation such that the discriminative information for subsets of classes are acquired in separate subspaces, which can help reduce the effect of class imbalance on the structure the deep feature space.}

{In the empirical study, the proposed framework showed advantages over the previous work on difficult problems with larger number of classes and/or smaller number of samples, and also maintained competitive performance given weak-labels which are not relevant to the task at hand.}

{In this paper, we limited the description of the proposed approach to handling one set of abstract-labels. However, it can naturally exploit multiple sets of labels by designating a subspace for each combination of labels.}

\ifBibtex
\bibliographystyle{splncs04}
\bibliography{/Users/ando/OneDrive/Bibliography/All}
\else

\fi

\end{document}